\pdfoutput=1
\documentclass[10pt,letterpaper]{article}

\usepackage{cogsci}
\usepackage[margin=1in]{geometry} 
\usepackage{microtype}
\usepackage{graphicx}
\usepackage{subfigure}
\usepackage{booktabs} 
\usepackage{nicefrac}       
\usepackage{microtype}      
\usepackage{float}
\usepackage{hyperref}
\hypersetup{colorlinks,allcolors=purple}
\usepackage{algorithm}
\usepackage{amsmath}
\usepackage{graphicx}
\usepackage[table,xcdraw]{xcolor}
\usepackage{gensymb}
\usepackage{stmaryrd}
\usepackage{amssymb}

\cogscifinalcopy 

\usepackage{pslatex}
\usepackage{apacite}

\title{Structured, flexible, and robust: benchmarking and improving large language models towards more human-like behavior in out-of-distribution reasoning tasks}
 
 \author{
    {\large \bf Katherine M. Collins}$^{1, 2\star \dagger}$, {\large \bf Catherine Wong}$^{2 \star}$, {\large \bf Jiahai Feng}$^2$,\\ 
    {\large \bf Megan Wei}$^2$, \and {\large \bf Joshua B. Tenenbaum}$^{2}$\\ 
    $^1$University of Cambridge,
    $^2$MIT\\
    $^{\dagger}$\texttt{kmc61@cam.ac.uk}}

\begin{document} 

\maketitle

\renewcommand{\thefootnote}{\fnsymbol{footnote}}
\footnotetext[1]{Contributed equally.}
\footnotetext[3]{Data and code for the project can be found at: \href{https://github.com/collinskatie/structured_flexible_and_robust}{https://github.com/collinskatie/structured\_flexible\_and\_robust}}

\begin{abstract}
Human language offers a powerful window into our thoughts -- we tell stories, give explanations, and express our beliefs and goals through words. Abundant evidence also suggests that language plays a developmental role in structuring our learning. Here, we ask: how much of human-like \textit{thinking} can be captured by learning statistical patterns in language alone? We first contribute a new challenge benchmark for comparing humans and distributional large language models (LLMs). Our benchmark contains two problem-solving domains (\textit{planning} and \textit{explanation} generation) and is designed to require generalization to new, out-of-distribution problems expressed in language. We find that humans are far more robust than LLMs on this benchmark. Next, we propose a hybrid \textit{Parse-and-Solve} model, which augments distributional LLMs with a structured symbolic reasoning module. We find that this model shows more robust adaptation to out-of-distribution planning problems, demonstrating the promise of hybrid AI models for more human-like reasoning.
\end{abstract}

\section{Introduction} \label{sec-introduction}

\noindent Language expresses the rich internal landscape of our thinking in a form that can be shared externally with others.  
We tell stories about real (\textit{what did I do today?}) and hypothetical (\textit{what would I do if I won the lottery?}) situations; give instructions for achieving goals ranging from the mundane (\textit{how do I put away the dishes?}) to the complex (\textit{how do I fix a carburetor?}); and propose explanations for both everyday events (\textit{why isn't the light bulb turning on?}) and novel observations (\textit{what's that strange beeping sound?}).  Learning language and learning \textit{from} language also play crucial roles in the development of children's thinking \shortcite{gopnik1997words,carey2009our,harris2018cognitive}.  But what, in computational terms, is the relationship between language and thought, and between learning language and learning to think? 

Classical theories draw a stark division between {\em thinking} as the manipulation of structured representations in an internal symbol system or language of thought (LOT) \cite{fodorLOT}, and {\em language} as a system of mappings between those representations and outwardly expressed forms (e.g., sounds, text). Under this view, learning language plays at best a supporting role in learning to think. Recently however, a new generation of statistical language learning systems in AI has put forth a serious challenge to this view. So-called \textit{large language models} (LLMs) \shortcite{gpt3,rae2021scaling} have demonstrated such striking success in realistic language production that they often appear to be ``thinking'' -- and yet they are driven solely by neural networks trained to predict the distribution of next words in long text sequences from very large corpora of human language. Other work has proposed using LLMs as a universal foundation for emulating many human reasoning abilities -- including capacities as diverse as \textit{physical reasoning} \shortcite{piqa}, \textit{task-level planning} \shortcite{sharma2021skill, huang2022language}, and even \textit{mathematical reasoning} \shortcite{cobbe2021training} -- simply by re-framing them as linguistic prediction.  Under this view, ``all you need is language'': learning to think requires little more than learning (the statistics of) language, or learning only the latent structure sufficient to produce the most probable next word in any linguistic context.  

In this paper, our goal is to critically assess how close modern LLMs come to actually learning to think, and to sketch out an alternative hybrid view of the language-thought interface that \textit{integrates} elements of the classical LOT and recent LLM paradigms.  In Part I, we describe a new, generic approach for constructing \textit{linguistic reasoning prompts} that measure flexible, creative thinking abilities in novel situations, as opposed to the ability to retrieve familiar patterns of thought for familiar situations.
We use an \textit{iterative constraint generation} paradigm that extends initial linguistic prompts using linguistic \textit{constraints} that restrict production of the most common human responses, forcing responses that require novel language production -- and, we argue, a greater degree of thinking. 
We compare LLMs to humans using this benchmark on two domains -- \textit{plan} and \textit{explanation} generation -- and find that humans both significantly outperform LLMs in general, and are comparatively more robust to prompts that extend beyond the standard distribution of human language. In Part II, we propose an alternative computational approach that leverages an LLM to map natural language into a space of structured programs, such that reasoning problems can be solved by powerful, scalable symbolic algorithms - rather than the purely neural form of end-to-end LLMs alone. We implement and demonstrate this model in a simplified synthetic language setting designed to emulate the \textit{planning domain} in Part I. Our results suggest that such hybrid approaches are a promising way forwards, albeit still rich with potential for future improvement.

\begin{figure*}[ht!]
  \begin{center}
  \includegraphics[width=0.99\linewidth]{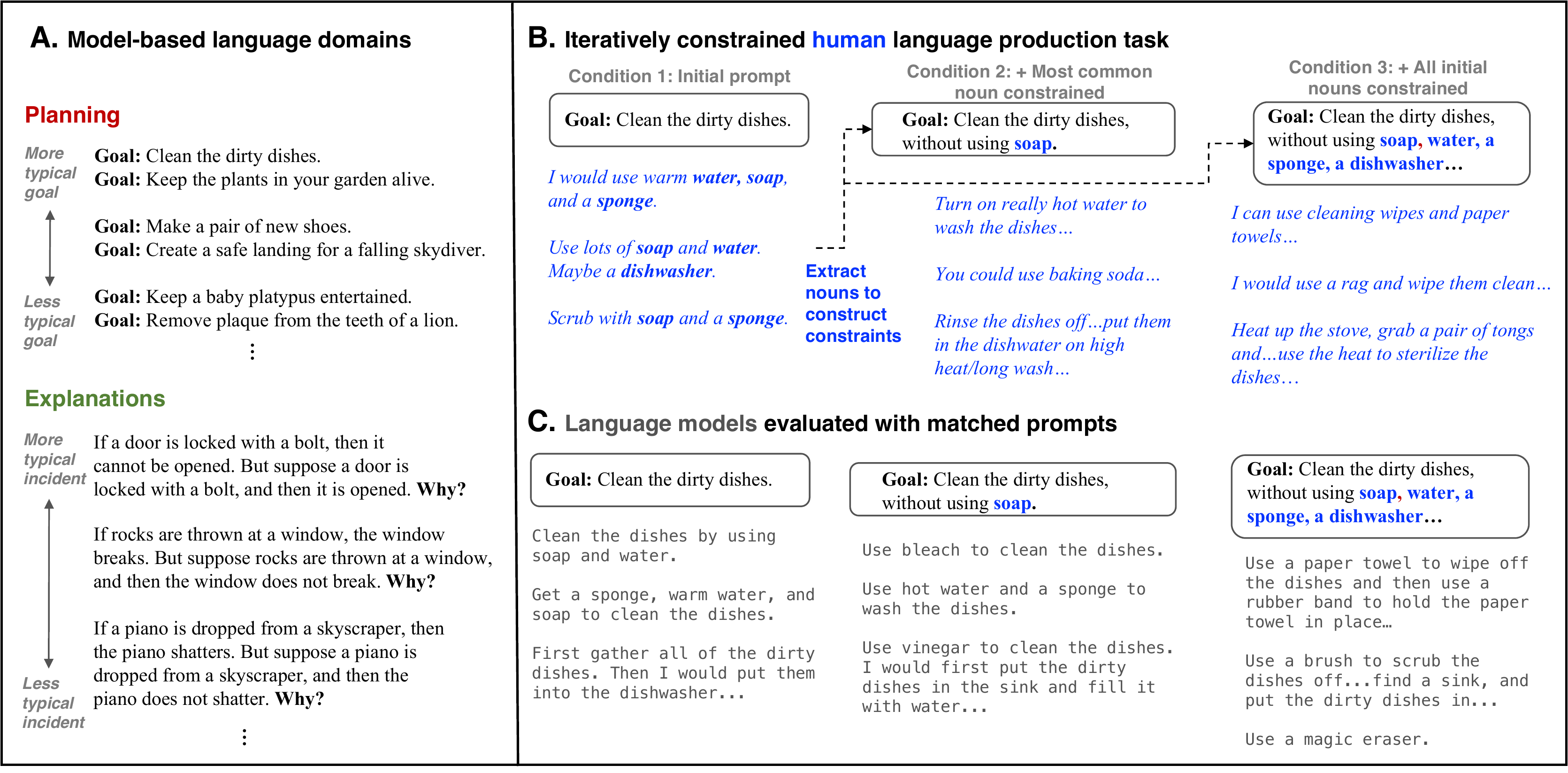}
  \caption{Iterative reasoning task overview. A) Sample goals and scenarios for the planning and explanation domains, respectively, illustrating the range of base typicality of our stimuli; B) Formation of constraints from human-generated language, where constraints are selected based on frequency, with sample \textcolor{blue}{human generations} (blue text) C) \textcolor{gray}{LLM-generations} (gray text) in response to the same prompts.}
  \label{fig:overview}
  \end{center}
  \end{figure*}

    

\section{Part I: Linguistic reasoning benchmark for humans and language models}
 
The first core motivation of this work is to evaluate the extent to which modeling the \textit{predictive distribution of language} actually captures the underlying \textit{reasoning} latent in human language. Towards this end, we propose a benchmark task (Fig. \ref{fig:overview}) based on two core reasoning abilities -- \textit{goal-based planning} and \textit{causal explanation} -- using an iterative design to challenge models which simply learn predictable responses from prior language.

\subsection{Methods}
We benchmark human and language model performance using a two-stage experimental design. In the first stage, an iterative \textit{human language production experiment} (Fig. \ref{fig:overview}B), we collect human responses on two domains (\textbf{planning} and \textbf{explanations}) under three progressively more challenging conditions: a baseline \textbf{initial prompt} condition using a collecting of linguistic reasoning prompts; and \textbf{two constrained conditions} which restrict the use of common answers to each prompt, in order to encourage participants to generate novel linguistic solutions. In the second stage, we evaluate a \textit{large language model} (LLM) on the same prompts, and collect responses by sampling from its predictive distribution.  We describe each stage in more detail below.

\paragraph{Human language production experiment}\label{iterativeReasoning} 

\paragraph{\textit{Participants}} 240 participants recruited from Prolific (2 domains x 3 conditions x 40 participants) completed the task. Base pay was \$15/hr, with a \$1 quality bonus.

\paragraph{\textit{Condition 1: initial reasoning prompts}} To measure baseline performance, our first reasoning condition elicits human responses to \textbf{initial prompts} (Fig. \ref{fig:overview}B, \textit{Condition 1}) on each grounding domain. We construct 28 \textit{goal prompts} for the planning domain (Fig. \ref{fig:overview}A, top), designed to elicit a concrete linguistic plan and to vary in their base typicality (eg. ranging from \textit{clean the dirty dishes} to \textit{get a sofa on the roof}). We also construct 28 \textit{causal event} prompts of varying typicality for the explanations domain (Fig. \ref{fig:overview}A, bottom), inspired by the ``unusual event" prompts in \shortcite{khemlani}: each event begins with an inciting cause and its usual consequence, then poses a counterfactual. 

Participants in this condition responded to a random batch (n=7) of prompts from a single domain, resulting in 10 unique responses per prompt. After responding to all prompts, we also ask participants to score \textit{base typicality} for each prompt of the goal (on planning) or inciting event (on explanations) using a 7-point Likert scale.

\paragraph{\textit{Condition 2 and 3: constrained reasoning prompts}}
In the subsequent conditions (Fig. \ref{fig:overview}B, \textit{Condition 2, 3}), we evaluate the human ability to flexibly generate more novel plans and explanations for the same initial prompts, by restricting their responses to prevent subjects from falling back on the most common solutions. Specifically, we use subject responses from Condition 1 to determine common (and likely highly \textit{predictable}) components of plans and explanations for each prompt. We construct linguistic constraints by extracting concrete \textit{nouns} from all responses to a given prompt (using an expert human tagger, who also lemmatizes and standarizes the form of each noun). We then extend each initial prompt in two more challenging conditions: in the \textbf{most common noun constrained} condition, we restrict responses which use the single most common noun; in the \textbf{all initial nouns constrained}, we restrict \textit{all} nouns which appear in the initial responses.

A new set of participants responded to a random batch (n=7) of prompts in a single domain and condition, again resulting in 10 unique responses per prompt and condition that reflect these linguistic constraints.

\paragraph{Language model matched production experiment}\label{llmGenPartI}
  
Our human experiment yields a series of linguistic prompts, in which individual goal and explanation prompts are extended across two more challenging conditions through linguistic \textit{constraints} that restrict the usage of the most common responses to each.

We use these same prompts to construct a benchmark language production task for our artificial language model. We evaluate our prompts on the state-of-the-art model \textit{GPT-3} \shortcite{gpt3}, using the \textit{few-shot prompting} technique introduced in \shortcite{gpt3} for generating predictive language for particular tasks. Specifically, we seed the model with a small number of \textit{examples} (n=12 goals, and n=15 explanations: the maximum number of examples the model allowed, based on token limits) pairing heldout prompts and human-generated text, then elicit generated responses for each prompt across all conditions.

To eliminate purely degenerate text, we also \textit{prescreen} the samples by asking human evaluators (N=370; recruited from Prolific) to score responses for surface language errors alone, and remove the lowest scoring responses. After screening, we collect a total of 20 LLM-generated responses for each prompt in each condition.

\paragraph{Blind comparative human evaluation}
Having collected human and LLM responses to the same linguistic prompts across all conditions, we now benchmark their relative performance using blind human evaluators (N=393; recruited from Prolific) asked to evaluate responses in a single domain and condition a 7-point Likert scale (1: worst; 7: best). Subjects rated responses for a random batch of \textit{prompts}, scoring a (randomly shuffled) set of human (n=10) and LLM (n=10) responses for each.

\subsection{Results}
\begin{figure*}[h!]
  \begin{center}
  \includegraphics[width=0.99\linewidth]{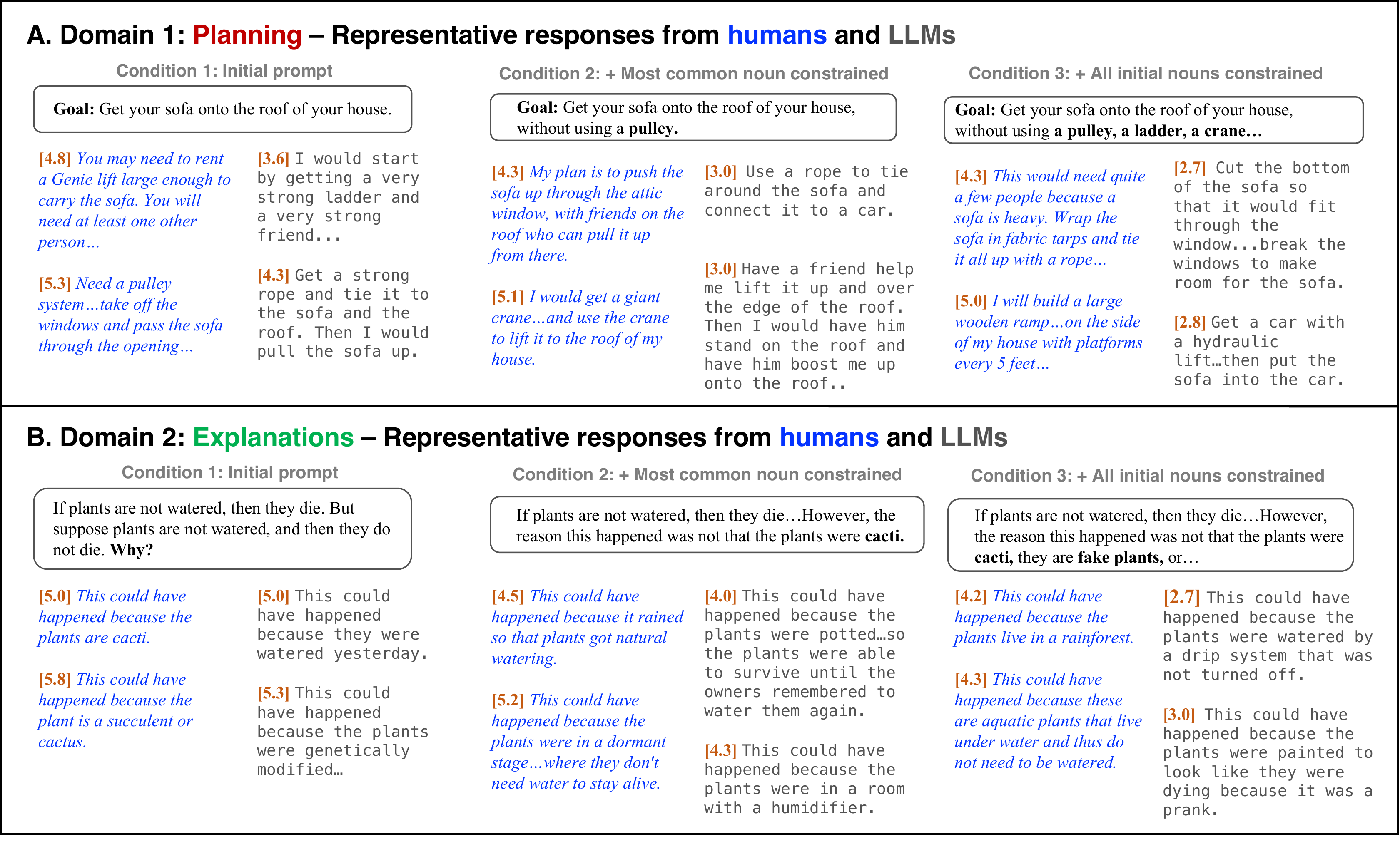}
  \caption{Representative plans (A) and explanations (B), per constraint condition, generated by \textcolor{blue}{humans} and an end-to-end \textcolor{gray}{LLM}. Average goodness rating, over the human evaluators for each generation, is shown in \textcolor{orange}{orange}.}
  \label{fig:languageGallery}
  \end{center}
  \end{figure*}
\begin{figure*}[h!]
  \begin{center}
  \includegraphics[width=0.99\linewidth]{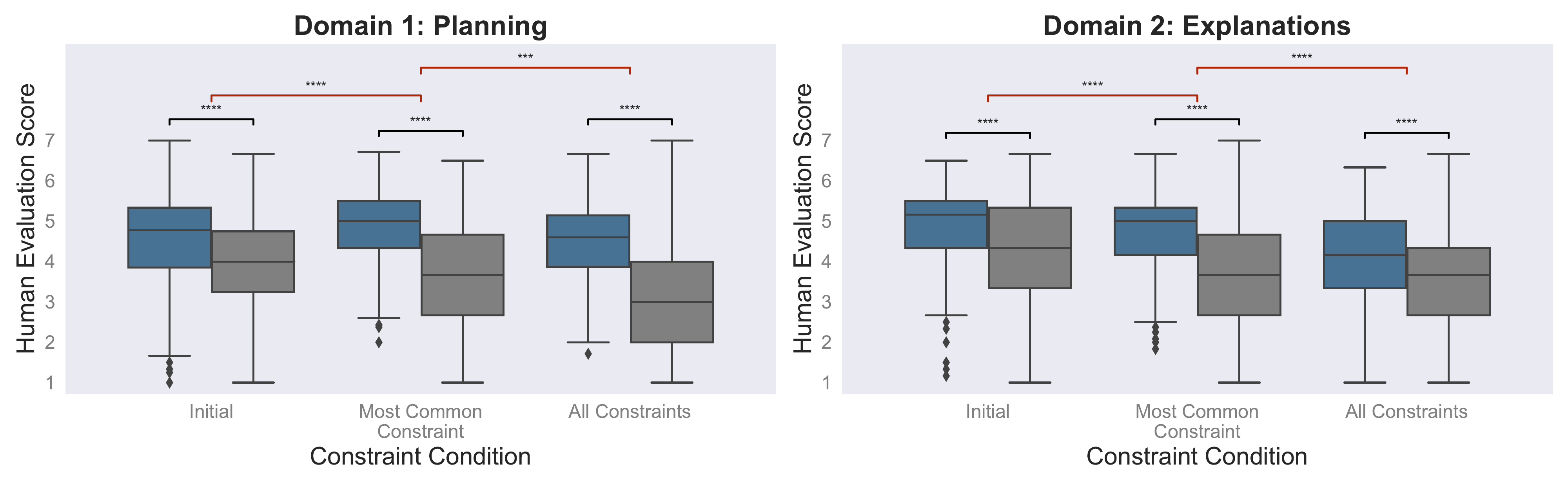}
  \caption{Mean overall goodness rating over plans (left) and explanations (right), show across all three constraint conditions. \textcolor{blue}{Humans} (blue boxes) significantly outperform the \textcolor{gray}{LLM} (gray boxes) in every condition (black, lower bars) and in successive pairwise conditions (red, upper bars).}
  \label{fig:modelResults}
  \end{center}
  \end{figure*}

Representative \textit{human} responses and \textit{language model} responses across both domains and conditions are depicted in Fig. \ref{fig:languageGallery}. To investigate comparative performance, we fit linear mixed effects regression (LMER) models predicting the human-evaluated score and use a corresponding likelihood ratio test (LRT) between an ablated model to determine the significance of the fixed effects. Fig. \ref{fig:modelResults} shows results of the blind human evaluation, and depicts statistical significance within and across conditions. 

\paragraph{People outperform the LLM within each reasoning condition} We first fit a LMER predicting the human evaluated score from the \textit{source} language generator (human or LLM), with random effects for the individual raters and prompts (syntax: \texttt{score $\sim$ source + (1 | rater\_id) + (1 | prompt)}). Our LRT finds that there is a significant effect ($p < 0.001$) of the language source (humans vs. LLM) in both domains and in each condition (\ref{fig:modelResults}, black indicators), humans outperform the LLM in \textit{every condition}, across both domains.

\paragraph{People are more robust to out-of-distribution prompts with constraints} We next consider our more central question: how well do language models perform specifically on our more \textbf{constrained} conditions, designed explicitly to force both humans and models to generate novel solutions to our underlying reasoning task? We expect humans to not only outperform language models in a direct comparison across individual prompts, but also to be comparatively more \textit{robust} to prompts which restrict highly predictable answers, and require responses beyond the distribution of standard human language.

An initial LMER with a fixed effect for the \textit{condition} (\textit{unconstrained}, \textit{most common constraint}, or \textit{many constraints}) suggests that both humans and LLMs are sensitive to the added constraints, though we find a strongly significant effect of condition on performance for LLMs ($p < 0.001$); and a weakly significant effect ($p=0.03$) for humans in the planning domain but strongly significant for explanations ($p < 0.001$).

However, a subsequent LMER with an interaction term for the language source (humans or LLMs) and condition (fit pairwise across each successive set of conditions) indicates that humans and LLMs are not \textit{equally} sensitive to constraints: we find strongly signficant interaction terms (Fig. \ref{fig:modelResults}, red) indicating that humans are more \textit{robust} to added constraints across each condition. This supports our central hypothesis: language models are increasingly poor at solving the underlying task once the prompts are constrained to restrict predictable responses.

\paragraph{People are more robust to \textit{goal} typicality} We also investigate whether another measure of linguistic predictability -- the atypicality of our base prompts -- also impacts LLM performance relative to humans. We fit a final LMER model with an interaction term for source and human typicality scores elicited in our initial experiment. Interestingly, we find a significant  interaction effect of typicality ($p < 0.001)$ for the \textit{planning} domain, but not for explanations. As assessing typicality for these prompts is more complex, further work (such as linguistic measures of prompt typicality) are necessary to better assess the explanations domains. This finding further supports our broader hypothesis: that LLMs are less robust to responding to out-of-distribution scenarios which pose novel, but solvable, planning problems.

\paragraph{Qualitative analysis of commonsense failures in LLM reasoning} Do large language models suffer from distinctively \textit{different} patterns of errors? An initial, qualitative examination suggests that large language models are particularly prone to errors indicating a more fundamental lack of ``common sense" understanding: of the underlying task, or the world knowledge required to solve it. A preliminary examination suggests that language models struggle particularly in generating coherent, realistic solutions for problems that require novel but concrete physical reasoning: as in the \textit{sofa on a roof} goals in Fig. \ref{fig:languageGallery}; or failures to understand \textit{color} (\textit{The carpet was white, so the blue dye did not show up}); \textit{water} (the \textit{grass is not made of water and so it does not absorb the water}); or gravity and \textit{material} (eg. someone failing to scrape their knees after falling in \textit{pants that were made of paper}).
Taken together, our reasoning experiment suggests that despite the surface plausibilty of their generated text, large language models generally struggle to emulate the latent reasoning that backs human responses -- once problems expressed in language require solutions beyond the standard, and most predictable, distribution of prior language, the apparent ``reasoning" abilities of these models deteriorate sharply.

\section{Part II: Integrating language with structured reasoning models}
Our results in Part I suggest that even very large language models may not capture the characteristic flexibility of human reasoning: they struggle to produce language reflecting novel computation over an underlying task.

\begin{figure*}[h!]
  \begin{center}
    \includegraphics[width=0.99\linewidth]{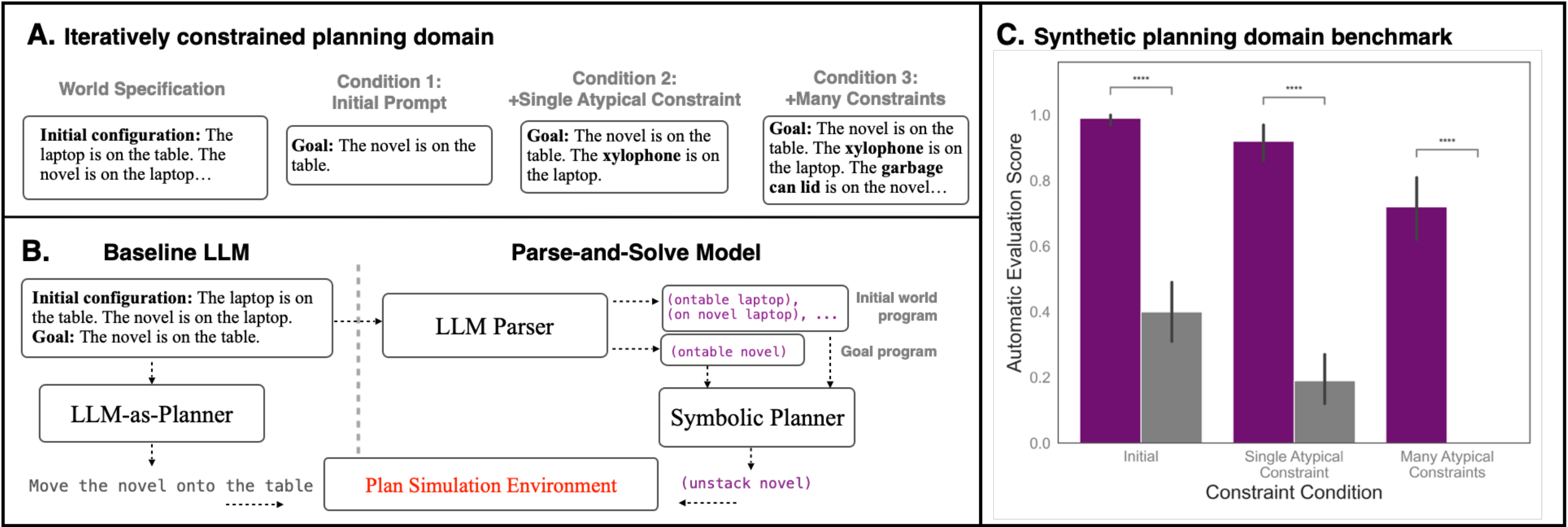}
  \caption{Simulated iterative planning task overview. A) Example progressively-constrained goal stimuli; B) Evaluation compares plans generated directly from an LLM (left) with plans generated from P+S (right); C) Success rate of \textcolor{violet}{P+S model} (purple) vs. \textcolor{gray}{LLM} (gray); P+S statistically significantly outperforms the LLM under each condition (black bars).}
  \label{fig:parsePlanModel}
  \end{center}
  \end{figure*}

Here, we propose an alternate computational approach for reasoning about problems posed in language. Rather than hoping to simulate latent computations (like planning) by directly predicting output language, we propose a simple (but demonstrative) \textit{parse-and-symbolic planner} (P+S) model which grounds language in an explicit ``language-of-thought" \cite{fodorLOT}: a formal \textit{program} expressing the meaning of the linguistic prompt, which interfaces with a symbolic computational solver (Fig. \ref{fig:parsePlanModel}B). 

\subsection{Simulated planning experiment}
We introduce a \textit{simulated planning domain} to benchmark our \textit{parse-and-symbolic planner} model against a standard LLM (here, GPT-Neo \cite{gpt-neo}), using a restricted set of prompts designed to emulate the core properties of the broader planning domain in Part I. We focus on \textit{planning} here for a straightforward metric of comparative performance: accuracy of our restricted plans can be evaluated directly on an explicit world model.

\paragraph{Initial and constrained synthetic planning prompts} As with Part I, our simulated experiment benchmarks model performance under three progressively more challenging conditions: responses to an \textbf{initial set} of linguistic goal prompts (Fig. \ref{fig:parsePlanModel}A, \textit{Condition 1}); and \textbf{two constrained conditions} which introduce new linguistic constraints over the initial goal (Fig. \ref{fig:parsePlanModel}B, \textit{Condition 2, 3}). As is obvious from Fig. \ref{fig:parsePlanModel}B, our conditions differ from Part I in one important respect: we extend our initial goals with \textit{positive constraints}, rather than the negative constraints in Part I. This format permits a more direct, albeit simplified, evaluation of the core task -- fully simulating \textit{restrictions} on initial resources would require modeling (and communicating) all possible alternative ways to achieve a goal in a simulated environment  -- while still requiring models to reason about complex, out-of-distribution language.

We generate initial and constrained goal prompts -- along with a linguistic \textit{initial condition} completely specifying the starting planning state for each prompt -- from a synthetic grammar over a simple \textit{object-stacking domain} \shortcite{gupta1992complexity}, in which each goal is a target stack of objects on a table (Fig. \ref{fig:parsePlanModel}). \textit{Initial prompts} involve goals with a single common household object; these are extended with both a \textbf{single constraint} and \textbf{many constraints} (n=4) that introduce additional, \textit{unusual} objects into the initial goal. In total, we sample n=100 initial goals and then sample constraints for both extended conditions.

\paragraph{Parse-and-solve model} Fig.  \ref{fig:parsePlanModel}B depicts a schematic of our parse-and-solve model, designed to disentangle language from the underlying computation required to solve planning tasks expressed in language. Our model integrates two distinct components. First, it \textit{parses} language into a formal \textit{program} representing the initial problem state and goal (using the PDDL planning language \shortcite{AICPub1821:1998}). For more direct comparison with a benchmark LLM, we also use a \textit{large language model} as our surface \textit{parser}: we use the Codex \shortcite{codex} model (a GPT-3 model fine-tuned on a joint distribution of language and symbolic programs), which can ``parse" language into programs using an analogous few-shot prompting technique (seeded with coupled examples of text and code). Unlike our comparison model, however, we employ distributional prediction only for a more constrained task: emulating the joint variation between a natural and formal language. 
The parsed \textit{programs} are passed to our model's second core component: a symbolic \textit{solver}, modeled with a search-based planner \shortcite{alkhazraji-et-al-zenodo2020} which attempts to generate a symbolic plan over a restricted set of actions (\textit{moving} objects from one location to the next) to solve the parsed goal.

\paragraph{Plan simulation environment}
Unlike in Part I, plans using the restricted space of actions in this domain can be simulated directly to assess accuracy. The P+S model outputs executable PDDL actions; the LLM-as-planner baseline outputs language which we reparse by inverting the synthetic grammar into PDDL actions. For both models, we mark unparseable or invalid plans as unsuccessful.

\subsection{Results} 
Analogous analyses to those in Part I (Fig. \ref{fig:parsePlanModel}C) -- measuring the comparative performance of our model with an LLM, as well as its \textit{robustness} to constraints -- suggest that our hybrid model, which uses predictive modeling only to transform language into a structured interface to an underlying symbolic planner, vastly improves its ability to adapt to complexly constrained goals.
\paragraph{Parse-and-solve model outperforms LLM} An LMER comparing our two models (P+S and LLM) finds a strongly significant difference in overall performance ($p < 0.001$; Fig. \ref{fig:parsePlanModel}C): indeed, the LLM solves \textit{none} of the problems in our most constrained condition.

\paragraph{Comparative robustness to constraints} Interestingly, a pairwise LMER testing for an interaction between source and condition does not find a significant interaction effect, suggesting that both models decline similarly in relative performance between conditions. One likely possibility is that this is an artifact of our restricted experiment size: the LLM simply can perform no worse in the final condition. However, these results could also suggest that the parsing approach we use here -- which employs distributional models to map language into programs -- may itself struggle to generalize; a hybrid \textit{parser}, which itself draws on more structured representations (like classical \textit{linguistic} grammars), might be better suited to parsing our most challenging compositional goals.

\section{Discussion} \label{sec-discussion}
Human language provides a richly structured window into how we think about the world. Our results, however, suggest that modeling the distribution of language alone may not be sufficient to capture the computations underlying planning, explanations, and other forms of reasoning which ground the language we produce. Instead, we propose an alternative approach: hybrid models which use distributional prediction to map language into structured formal representations of meaning that interface directly with structured symbolic algorithms \shortcite{dreamcoder, laps, sys1sys2}. Our contributions here leave much open for future work: to more systematically characterize regimes under which simply producing probable language closely approximates, and \textit{deviates}, from human reasoning, and go beyond the simple demonstration model we have provided towards broader-coverage models for more realistic reasoning domains. 

An important next step will be building on the qualitative analyses in Part I to disentangle the many factors (e.g., \textit{accuracy}, \textit{semantic coherence}, and \textit{concision}) that may separate human performance from purely predictive responses. In tandem, the hybrid model we propose here offers a promising, albeit highly restricted, step towards emulating human-like reasoning over language. How do we \textit{learn} the structured world models, or even sophisticated planning algorithms, that our simple model builds upon? Our core modeling approach suggests a path towards these more fundamental learning problems: using language to construct, or guide discovery, of \textit{programs} which represent novel environments, actions, and even algorithms for operating over such worlds.

\hfill
\bibliographystyle{apacite}

\setlength{\bibleftmargin}{.125in}
\setlength{\bibindent}{-\bibleftmargin}

\section{Acknowledgments}
We thank Laura Schulz, Junyi Chu, Alex Lew, Joao Loula Guimaraes de Campos, Max Nye, and the rest of the GPS Community for many thrilling, inspiring conversations, as well as practical advice with our project. We are also deeply grateful for Pratyusha Sharma, Jacob Andreas, and Noa Korneev for their thoughtful suggestions and support. We thank Yoni Friedman for his fantastic help with human annotator recruitment. Additionally, we thank the OpenAI team for increasing our quota to enable us to run more GPT-3 rollouts for Part I, and our Anonymous Reviewers for useful comments. \\\\
KMC is supported by a Marshall Scholarship and conducted work on the project under a Goldwater Scholarship. CW and JBT are supported by AFOSR\#FA9550-19-1-0269, the MIT Quest for Intelligence, the MIT-IBM Watson AI Lab, ONR Science of AI, and DARPA Machine Common Sense.

\bibliography{structured_thought_refs}

\begin{thebibliography}{}

\bibitem [\protect \citeauthoryear {%
Alkhazraji%
\ \protect \BOthers {.}}{%
Alkhazraji%
\ \protect \BOthers {.}}{%
{\protect \APACyear {2020}}%
}]{%
alkhazraji-et-al-zenodo2020}
\APACinsertmetastar {%
alkhazraji-et-al-zenodo2020}%
\begin{APACrefauthors}%
Alkhazraji, Y.%
, Frorath, M.%
, Gr{\"u}tzner, M.%
, Helmert, M.%
, Liebetraut, T.%
, Mattm{\"u}ller, R.%
\BDBL {}W{\"u}lfing, J.%
\end{APACrefauthors}%
\unskip\
\newblock
\APACrefYearMonthDay{2020}{}{}.
\newblock
\APACrefbtitle {Pyperplan.} {Pyperplan.}
\newblock
\APAChowpublished {\url{https://doi.org/10.5281/zenodo.3700819}}.
\newblock
\APACaddressPublisher{}{Zenodo}.
\newblock
\begin{APACrefURL} \url{https://doi.org/10.5281/zenodo.3700819}
  \end{APACrefURL}
\newblock
\begin{APACrefDOI} \doi{10.5281/zenodo.3700819} \end{APACrefDOI}
\PrintBackRefs{\CurrentBib}

\bibitem [\protect \citeauthoryear {%
Bisk%
, Zellers%
, Bras%
, Gao%
\BCBL {}\ \BBA {} Choi%
}{%
Bisk%
\ \protect \BOthers {.}}{%
{\protect \APACyear {2019}}%
}]{%
piqa}
\APACinsertmetastar {%
piqa}%
\begin{APACrefauthors}%
Bisk, Y.%
, Zellers, R.%
, Bras, R\BPBI L.%
, Gao, J.%
\BCBL {}\ \BBA {} Choi, Y.%
\end{APACrefauthors}%
\unskip\
\newblock
\APACrefYearMonthDay{2019}{}{}.
\newblock
{\BBOQ}\APACrefatitle {{PIQA:} Reasoning about Physical Commonsense in Natural
  Language} {{PIQA:} reasoning about physical commonsense in natural
  language}.{\BBCQ}
\newblock
\APACjournalVolNumPages{CoRR}{abs/1911.11641}{}{}.
\newblock
\begin{APACrefURL} \url{http://arxiv.org/abs/1911.11641} \end{APACrefURL}
\PrintBackRefs{\CurrentBib}

\bibitem [\protect \citeauthoryear {%
Black%
, Gao%
, Wang%
, Leahy%
\BCBL {}\ \BBA {} Biderman%
}{%
Black%
\ \protect \BOthers {.}}{%
{\protect \APACyear {2021}}%
}]{%
gpt-neo}
\APACinsertmetastar {%
gpt-neo}%
\begin{APACrefauthors}%
Black, S.%
, Gao, L.%
, Wang, P.%
, Leahy, C.%
\BCBL {}\ \BBA {} Biderman, S.%
\end{APACrefauthors}%
\unskip\
\newblock
\APACrefYearMonthDay{2021}{{\APACmonth{03}}}{}.
\newblock
\APACrefbtitle {{GPT-Neo: Large Scale Autoregressive Language Modeling with
  Mesh-Tensorflow}.} {{GPT-Neo: Large Scale Autoregressive Language Modeling
  with Mesh-Tensorflow}.}
\newblock
\APACaddressPublisher{}{Zenodo}.
\newblock
\begin{APACrefURL} \url{https://doi.org/10.5281/zenodo.5297715}
  \end{APACrefURL}
\newblock
\APACrefnote{{If you use this software, please cite it using these metadata.}}
\newblock
\begin{APACrefDOI} \doi{10.5281/zenodo.5297715} \end{APACrefDOI}
\PrintBackRefs{\CurrentBib}

\bibitem [\protect \citeauthoryear {%
Brown%
\ \protect \BOthers {.}}{%
Brown%
\ \protect \BOthers {.}}{%
{\protect \APACyear {2020}}%
}]{%
gpt3}
\APACinsertmetastar {%
gpt3}%
\begin{APACrefauthors}%
Brown, T\BPBI B.%
, Mann, B.%
, Ryder, N.%
, Subbiah, M.%
, Kaplan, J.%
, Dhariwal, P.%
\BDBL {}Amodei, D.%
\end{APACrefauthors}%
\unskip\
\newblock
\APACrefYearMonthDay{2020}{}{}.
\newblock
{\BBOQ}\APACrefatitle {Language Models are Few-Shot Learners} {Language models
  are few-shot learners}.{\BBCQ}
\newblock
\APACjournalVolNumPages{CoRR}{abs/2005.14165}{}{}.
\newblock
\begin{APACrefURL} \url{https://arxiv.org/abs/2005.14165} \end{APACrefURL}
\PrintBackRefs{\CurrentBib}

\bibitem [\protect \citeauthoryear {%
Carey%
}{%
Carey%
}{%
{\protect \APACyear {2009}}%
}]{%
carey2009our}
\APACinsertmetastar {%
carey2009our}%
\begin{APACrefauthors}%
Carey, S.%
\end{APACrefauthors}%
\unskip\
\newblock
\APACrefYearMonthDay{2009}{}{}.
\newblock
{\BBOQ}\APACrefatitle {Where our number concepts come from} {Where our number
  concepts come from}.{\BBCQ}
\newblock
\APACjournalVolNumPages{The Journal of philosophy}{106}{4}{220}.
\PrintBackRefs{\CurrentBib}

\bibitem [\protect \citeauthoryear {%
Chen%
\ \protect \BOthers {.}}{%
Chen%
\ \protect \BOthers {.}}{%
{\protect \APACyear {2021}}%
}]{%
codex}
\APACinsertmetastar {%
codex}%
\begin{APACrefauthors}%
Chen, M.%
, Tworek, J.%
, Jun, H.%
, Yuan, Q.%
, de Oliveira~Pinto, H\BPBI P.%
, Kaplan, J.%
\BDBL {}Zaremba, W.%
\end{APACrefauthors}%
\unskip\
\newblock
\APACrefYearMonthDay{2021}{}{}.
\newblock
{\BBOQ}\APACrefatitle {Evaluating Large Language Models Trained on Code}
  {Evaluating large language models trained on code}.{\BBCQ}
\newblock
\APACjournalVolNumPages{CoRR}{abs/2107.03374}{}{}.
\newblock
\begin{APACrefURL} \url{https://arxiv.org/abs/2107.03374} \end{APACrefURL}
\PrintBackRefs{\CurrentBib}

\bibitem [\protect \citeauthoryear {%
Cobbe%
\ \protect \BOthers {.}}{%
Cobbe%
\ \protect \BOthers {.}}{%
{\protect \APACyear {2021}}%
}]{%
cobbe2021training}
\APACinsertmetastar {%
cobbe2021training}%
\begin{APACrefauthors}%
Cobbe, K.%
, Kosaraju, V.%
, Bavarian, M.%
, Hilton, J.%
, Nakano, R.%
, Hesse, C.%
\BCBL {}\ \BBA {} Schulman, J.%
\end{APACrefauthors}%
\unskip\
\newblock
\APACrefYearMonthDay{2021}{}{}.
\newblock
{\BBOQ}\APACrefatitle {Training verifiers to solve math word problems}
  {Training verifiers to solve math word problems}.{\BBCQ}
\newblock
\APACjournalVolNumPages{arXiv preprint arXiv:2110.14168}{}{}{}.
\PrintBackRefs{\CurrentBib}

\bibitem [\protect \citeauthoryear {%
Ellis%
\ \protect \BOthers {.}}{%
Ellis%
\ \protect \BOthers {.}}{%
{\protect \APACyear {2020}}%
}]{%
dreamcoder}
\APACinsertmetastar {%
dreamcoder}%
\begin{APACrefauthors}%
Ellis, K.%
, Wong, C.%
, Nye, M\BPBI I.%
, Sabl{\'{e}}{-}Meyer, M.%
, Cary, L.%
, Morales, L.%
\BDBL {}Tenenbaum, J\BPBI B.%
\end{APACrefauthors}%
\unskip\
\newblock
\APACrefYearMonthDay{2020}{}{}.
\newblock
{\BBOQ}\APACrefatitle {DreamCoder: Growing generalizable, interpretable
  knowledge with wake-sleep Bayesian program learning} {Dreamcoder: Growing
  generalizable, interpretable knowledge with wake-sleep bayesian program
  learning}.{\BBCQ}
\newblock
\APACjournalVolNumPages{CoRR}{abs/2006.08381}{}{}.
\newblock
\begin{APACrefURL} \url{https://arxiv.org/abs/2006.08381} \end{APACrefURL}
\PrintBackRefs{\CurrentBib}

\bibitem [\protect \citeauthoryear {%
Fodor%
}{%
Fodor%
}{%
{\protect \APACyear {1975}}%
}]{%
fodorLOT}
\APACinsertmetastar {%
fodorLOT}%
\begin{APACrefauthors}%
Fodor, J\BPBI A.%
\end{APACrefauthors}%
\unskip\
\newblock
\APACrefYear{1975}.
\newblock
\APACrefbtitle {The Language of Thought} {The language of thought}.
\newblock
\APACaddressPublisher{}{Harvard University Press}.
\PrintBackRefs{\CurrentBib}

\bibitem [\protect \citeauthoryear {%
Gopnik%
\ \BBA {} Meltzoff%
}{%
Gopnik%
\ \BBA {} Meltzoff%
}{%
{\protect \APACyear {1997}}%
}]{%
gopnik1997words}
\APACinsertmetastar {%
gopnik1997words}%
\begin{APACrefauthors}%
Gopnik, A.%
\BCBT {}\ \BBA {} Meltzoff, A\BPBI N.%
\end{APACrefauthors}%
\unskip\
\newblock
\APACrefYear{1997}.
\newblock
\APACrefbtitle {Words, thoughts, and theories} {Words, thoughts, and theories}.
\newblock
\APACaddressPublisher{}{Mit Press}.
\PrintBackRefs{\CurrentBib}

\bibitem [\protect \citeauthoryear {%
Gupta%
\ \BBA {} Nau%
}{%
Gupta%
\ \BBA {} Nau%
}{%
{\protect \APACyear {1992}}%
}]{%
gupta1992complexity}
\APACinsertmetastar {%
gupta1992complexity}%
\begin{APACrefauthors}%
Gupta, N.%
\BCBT {}\ \BBA {} Nau, D\BPBI S.%
\end{APACrefauthors}%
\unskip\
\newblock
\APACrefYearMonthDay{1992}{}{}.
\newblock
{\BBOQ}\APACrefatitle {On the complexity of blocks-world planning} {On the
  complexity of blocks-world planning}.{\BBCQ}
\newblock
\APACjournalVolNumPages{Artificial Intelligence}{56}{2-3}{223--254}.
\PrintBackRefs{\CurrentBib}

\bibitem [\protect \citeauthoryear {%
Harris%
, Koenig%
, Corriveau%
\BCBL {}\ \BBA {} Jaswal%
}{%
Harris%
\ \protect \BOthers {.}}{%
{\protect \APACyear {2018}}%
}]{%
harris2018cognitive}
\APACinsertmetastar {%
harris2018cognitive}%
\begin{APACrefauthors}%
Harris, P\BPBI L.%
, Koenig, M\BPBI A.%
, Corriveau, K\BPBI H.%
\BCBL {}\ \BBA {} Jaswal, V\BPBI K.%
\end{APACrefauthors}%
\unskip\
\newblock
\APACrefYearMonthDay{2018}{}{}.
\newblock
{\BBOQ}\APACrefatitle {Cognitive foundations of learning from testimony}
  {Cognitive foundations of learning from testimony}.{\BBCQ}
\newblock
\APACjournalVolNumPages{Annual Review of Psychology}{69}{}{251--273}.
\PrintBackRefs{\CurrentBib}

\bibitem [\protect \citeauthoryear {%
Huang%
, Abbeel%
, Pathak%
\BCBL {}\ \BBA {} Mordatch%
}{%
Huang%
\ \protect \BOthers {.}}{%
{\protect \APACyear {2022}}%
}]{%
huang2022language}
\APACinsertmetastar {%
huang2022language}%
\begin{APACrefauthors}%
Huang, W.%
, Abbeel, P.%
, Pathak, D.%
\BCBL {}\ \BBA {} Mordatch, I.%
\end{APACrefauthors}%
\unskip\
\newblock
\APACrefYearMonthDay{2022}{}{}.
\newblock
\APACrefbtitle {Language Models as Zero-Shot Planners: Extracting Actionable
  Knowledge for Embodied Agents.} {Language models as zero-shot planners:
  Extracting actionable knowledge for embodied agents.}
\PrintBackRefs{\CurrentBib}

\bibitem [\protect \citeauthoryear {%
Korman%
\ \BBA {} Khemlani%
}{%
Korman%
\ \BBA {} Khemlani%
}{%
{\protect \APACyear {2020}}%
}]{%
khemlani}
\APACinsertmetastar {%
khemlani}%
\begin{APACrefauthors}%
Korman, J.%
\BCBT {}\ \BBA {} Khemlani, S.%
\end{APACrefauthors}%
\unskip\
\newblock
\APACrefYearMonthDay{2020}{}{}.
\newblock
{\BBOQ}\APACrefatitle {Explanatory completeness} {Explanatory
  completeness}.{\BBCQ}
\newblock
\APACjournalVolNumPages{Acta Psychologica}{209}{}{103139}.
\newblock
\begin{APACrefURL}
  \url{https://www.sciencedirect.com/science/article/pii/S0001691819303531}
  \end{APACrefURL}
\newblock
\begin{APACrefDOI} \doi{https://doi.org/10.1016/j.actpsy.2020.103139}
  \end{APACrefDOI}
\PrintBackRefs{\CurrentBib}

\bibitem [\protect \citeauthoryear {%
McDermott%
\ \protect \BOthers {.}}{%
McDermott%
\ \protect \BOthers {.}}{%
{\protect \APACyear {1998}}%
}]{%
AICPub1821:1998}
\APACinsertmetastar {%
AICPub1821:1998}%
\begin{APACrefauthors}%
McDermott, D.%
, Ghallab, M.%
, Howe, A.%
, Knoblock, C.%
, Ram, A.%
, Veloso, M.%
\BDBL {}Wilkins, D.%
\end{APACrefauthors}%
\unskip\
\newblock
\APACrefYearMonthDay{1998}{}{}.
\newblock
\APACrefbtitle {PDDL - The Planning Domain Definition Language} {Pddl - the
  planning domain definition language}\ \APACbVolEdTR{}{\BTR{}\ \BNUM\
  TR-98-003}.
\newblock
\APACaddressInstitution{}{Yale Center for Computational Vision and Control,}.
\PrintBackRefs{\CurrentBib}

\bibitem [\protect \citeauthoryear {%
Nye%
, Tessler%
, Tenenbaum%
\BCBL {}\ \BBA {} Lake%
}{%
Nye%
\ \protect \BOthers {.}}{%
{\protect \APACyear {2021}}%
}]{%
sys1sys2}
\APACinsertmetastar {%
sys1sys2}%
\begin{APACrefauthors}%
Nye, M\BPBI I.%
, Tessler, M\BPBI H.%
, Tenenbaum, J\BPBI B.%
\BCBL {}\ \BBA {} Lake, B\BPBI M.%
\end{APACrefauthors}%
\unskip\
\newblock
\APACrefYearMonthDay{2021}{}{}.
\newblock
{\BBOQ}\APACrefatitle {Improving Coherence and Consistency in Neural Sequence
  Models with Dual-System, Neuro-Symbolic Reasoning} {Improving coherence and
  consistency in neural sequence models with dual-system, neuro-symbolic
  reasoning}.{\BBCQ}
\newblock
\APACjournalVolNumPages{CoRR}{abs/2107.02794}{}{}.
\newblock
\begin{APACrefURL} \url{https://arxiv.org/abs/2107.02794} \end{APACrefURL}
\PrintBackRefs{\CurrentBib}

\bibitem [\protect \citeauthoryear {%
Rae%
\ \protect \BOthers {.}}{%
Rae%
\ \protect \BOthers {.}}{%
{\protect \APACyear {2021}}%
}]{%
rae2021scaling}
\APACinsertmetastar {%
rae2021scaling}%
\begin{APACrefauthors}%
Rae, J\BPBI W.%
, Borgeaud, S.%
, Cai, T.%
, Millican, K.%
, Hoffmann, J.%
, Song, F.%
\BDBL {}others%
\end{APACrefauthors}%
\unskip\
\newblock
\APACrefYearMonthDay{2021}{}{}.
\newblock
{\BBOQ}\APACrefatitle {Scaling language models: Methods, analysis \& insights
  from training gopher} {Scaling language models: Methods, analysis \& insights
  from training gopher}.{\BBCQ}
\newblock
\APACjournalVolNumPages{arXiv preprint arXiv:2112.11446}{}{}{}.
\PrintBackRefs{\CurrentBib}

\bibitem [\protect \citeauthoryear {%
Sharma%
, Torralba%
\BCBL {}\ \BBA {} Andreas%
}{%
Sharma%
\ \protect \BOthers {.}}{%
{\protect \APACyear {2021}}%
}]{%
sharma2021skill}
\APACinsertmetastar {%
sharma2021skill}%
\begin{APACrefauthors}%
Sharma, P.%
, Torralba, A.%
\BCBL {}\ \BBA {} Andreas, J.%
\end{APACrefauthors}%
\unskip\
\newblock
\APACrefYearMonthDay{2021}{}{}.
\newblock
\APACrefbtitle {Skill Induction and Planning with Latent Language.} {Skill
  induction and planning with latent language.}
\PrintBackRefs{\CurrentBib}

\bibitem [\protect \citeauthoryear {%
Wong%
, Ellis%
, Tenenbaum%
\BCBL {}\ \BBA {} Andreas%
}{%
Wong%
\ \protect \BOthers {.}}{%
{\protect \APACyear {2021}}%
}]{%
laps}
\APACinsertmetastar {%
laps}%
\begin{APACrefauthors}%
Wong, C.%
, Ellis, K.%
, Tenenbaum, J\BPBI B.%
\BCBL {}\ \BBA {} Andreas, J.%
\end{APACrefauthors}%
\unskip\
\newblock
\APACrefYearMonthDay{2021}{}{}.
\newblock
{\BBOQ}\APACrefatitle {Leveraging Language to Learn Program Abstractions and
  Search Heuristics} {Leveraging language to learn program abstractions and
  search heuristics}.{\BBCQ}
\newblock
\APACjournalVolNumPages{CoRR}{abs/2106.11053}{}{}.
\newblock
\begin{APACrefURL} \url{https://arxiv.org/abs/2106.11053} \end{APACrefURL}
\PrintBackRefs{\CurrentBib}

\end{thebibliography}


\begin{thebibliography}{}

\bibitem [\protect \citeauthoryear {%
Black%
, Gao%
, Wang%
, Leahy%
\BCBL {}\ \BBA {} Biderman%
}{%
Black%
\ \protect \BOthers {.}}{%
{\protect \APACyear {2021}}%
}]{%
gpt-neo}
\APACinsertmetastar {%
gpt-neo}%
\begin{APACrefauthors}%
Black, S.%
, Gao, L.%
, Wang, P.%
, Leahy, C.%
\BCBL {}\ \BBA {} Biderman, S.%
\end{APACrefauthors}%
\unskip\
\newblock
\APACrefYearMonthDay{2021}{{\APACmonth{03}}}{}.
\newblock
\APACrefbtitle {{GPT-Neo: Large Scale Autoregressive Language Modeling with
  Mesh-Tensorflow}.} {{GPT-Neo: Large Scale Autoregressive Language Modeling
  with Mesh-Tensorflow}.}
\newblock
\APACaddressPublisher{}{Zenodo}.
\newblock
\begin{APACrefURL} \url{https://doi.org/10.5281/zenodo.5297715}
  \end{APACrefURL}
\newblock
\APACrefnote{{If you use this software, please cite it using these metadata.}}
\newblock
\begin{APACrefDOI} \doi{10.5281/zenodo.5297715} \end{APACrefDOI}
\PrintBackRefs{\CurrentBib}

\bibitem [\protect \citeauthoryear {%
Brown%
\ \protect \BOthers {.}}{%
Brown%
\ \protect \BOthers {.}}{%
{\protect \APACyear {2020}}%
}]{%
gpt3}
\APACinsertmetastar {%
gpt3}%
\begin{APACrefauthors}%
Brown, T\BPBI B.%
, Mann, B.%
, Ryder, N.%
, Subbiah, M.%
, Kaplan, J.%
, Dhariwal, P.%
\BDBL {}Amodei, D.%
\end{APACrefauthors}%
\unskip\
\newblock
\APACrefYearMonthDay{2020}{}{}.
\newblock
{\BBOQ}\APACrefatitle {Language Models are Few-Shot Learners} {Language models
  are few-shot learners}.{\BBCQ}
\newblock
\APACjournalVolNumPages{CoRR}{abs/2005.14165}{}{}.
\newblock
\begin{APACrefURL} \url{https://arxiv.org/abs/2005.14165} \end{APACrefURL}
\PrintBackRefs{\CurrentBib}

\bibitem [\protect \citeauthoryear {%
Chen%
\ \protect \BOthers {.}}{%
Chen%
\ \protect \BOthers {.}}{%
{\protect \APACyear {2021}}%
}]{%
codex}
\APACinsertmetastar {%
codex}%
\begin{APACrefauthors}%
Chen, M.%
, Tworek, J.%
, Jun, H.%
, Yuan, Q.%
, de Oliveira~Pinto, H\BPBI P.%
, Kaplan, J.%
\BDBL {}Zaremba, W.%
\end{APACrefauthors}%
\unskip\
\newblock
\APACrefYearMonthDay{2021}{}{}.
\newblock
{\BBOQ}\APACrefatitle {Evaluating Large Language Models Trained on Code}
  {Evaluating large language models trained on code}.{\BBCQ}
\newblock
\APACjournalVolNumPages{CoRR}{abs/2107.03374}{}{}.
\newblock
\begin{APACrefURL} \url{https://arxiv.org/abs/2107.03374} \end{APACrefURL}
\PrintBackRefs{\CurrentBib}

\end{thebibliography}

\end{document}


\maketitle

Here, we provide additional details on the human experiments, models, stimuli, and analyses described in Parts I and II of our main text. Further details can be found in \href{https://github.com/collinskatie/structured_flexible_and_robust}{our repository}. A pre-registration logged for Part I of our study can be found at \href{https://archive.org/details/osf-registrations-cy72b-v1}{here}. 

\section{S1. Part I Supplemental Details} \label{sec:s_part_i}

We first expand on the experimental design and results discussed in \textit{Part I: Linguistic reasoning benchmark for humans and language models}. Code for these experiments can be found in the \href{https://github.com/collinskatie/structured_flexible_and_robust/tree/main/Part_I}{``Part\_I'' directory} of {our repository}. 

\subsection{Creation of constraints} Constraints used in Condition 2 and 3 per domain were constructed by an expert human tagger from the plans and explanations produced by humans in Condition 1. The human tagger extracted all concrete nouns mentioned in the human generations. We allow the expert tagger to collapse multiple semantically identical phrases into a single noun, based on prior world knowledge; e.g., collapsing ``put a life net'', ``fly an outstretched net'', and ``allow them to land in the net'', into a single phrase ``the net.''

In the explanation domain, constraints were constructed with respect to the full set of $N=10$ unconstrained examples. In the planning domain, plans were generated in stages. In the first stage, approximately $4-7$ humans generated plans per goal, and in the second stage, additional humans were recruited to ensure at least $10$ saw each plan. Constraints for the planning domain were constructed only with respect to this first batch.  

\subsection{Human experiments}

We now provide additional details on the three flavors of human experiments run in this work: 1) generation of plans and explanations, 2) rating humanness of LLM-generated language, and 3) rating the overall goodness of mixed human- and LLM-generated language. In all experiments, participants were recruited from \href{https://www.prolific.co/}{Prolific} via \href{https://www.cognition.run/}{Cognition}. Participants were based in the United States and had to be at least $18$ years old and speak English. Participants were not allowed to partake in more than one flavor of experiment. 

\paragraph{Human language production and typicality rating} Language production tasks were designed to last $30$ minutes. Participants were asked to provide plans or explanations for seven goals or scenarios within a single condition.

Participants in Condition 1 (the initial prompt without constraints), also were asked to provide typicality ratings for all goals or scenarios that they had seen. All such ratings were conducted \textit{after} all seven language production tasks were completed. For the planning domain, participants were asked to mark on a 1-7 Likert scale ``How frequently you think people try to achieve each goal,'' where 1 = ``Most people do this on a daily basis.'' and 7 = ``I don’t see how this is even possible to do or try to do.'' Participants in the explanation domain rated prompts along two dimensions of typicality: 1) the base rate of the incident, X (e.g., ``How frequently do you think someone observes this event in the actual world?''), and 2) the rate of the effect, Y, given the cause-event X (e.g., ``Assuming that this initial event happens...how frequently do you think this results?''). 

\paragraph{Rating humanness} To give the LLM the fairest comparison, we employ human annotators to pre-screen, or filter, LLM generations based on their ``humanness.'' Annotators were asked to rate the likelihood that the \textit{language} could have been generated by a person on a 1-7 Likert scale (``How plausible is it that a human would have generated the language of the explanation/plan?'') with 1 being ``completely implausible'' and 7 being ``completely plausible''. Each rater saw a random subset of between 42-45 generations from within the same domain and associated constraint condition. The subset was randomly selected over the entire space of goals or scenarios, with the exception that a pilot check of the experiment was run for the planning domain only, where participants saw plans randomized over only three goals. We retained only explanations and plans which surpassed a score of $2$ to remove degenerate responses. 


\paragraph{Blind comparative human evaluation} Each participant rated language generated within the same domain, and within the same condition in said domain. A subset of $10$ of the $20$ LLM-generated plans or explanations was shuffled with all $10$ human plans. We divided the $20$ LLM generations into two sets of $10$ in advance. Each participant hence saw $20$ total ($10$ human, $10$ LLM) generated plans or explanations -- depending on the domain assigned -- for $3$ separate goals or scenarios, within the same condition. Participants had a $15$ second break after rating all language for a single goal or scenario, before continuing to the next. In the planning domain, annotators were asked: "How good is this plan overall? Assign it a single score that summarizes how good it is for this goal," and in the explanation domain: ``How good is this explanation overall? Assign it a single score that summarizes how good it is for this scenario.'' Ratings were marked on a 7-point Likert scale as noted in Part I. 

\subsection{Prompting GPT-3}

We next expand on the prompting regime used to glean plans and explanations from GPT-3 \cite{gpt3}. As an example, to generate plans for the goal ``Cool down in a record-breaking heat wave, without using an air conditioner,'' we sample $N=15$ of the remaining $27$ goals \textit{within the same condition} -- and for each of this $N$ goals, we randomly sample one of the $10$ human-generated plan written to achieve that goal. 

We concatenate these \textit{goal:plan} pairs in the following format: 

\begin{spverbatim}
"Goal: Help your local town mayor win re-election, without using billboards.\nPlan: \"Organize a team of people to campaign door-to-door. We can also print flyers to pass out on the streets and put on cars. A Facebook ad would be useful, as well as a radio interview if we can set one up. Finally, booking a debate with his opponent would help.\"\nGoal: Build a float to dazzle the crowd at the Macy's Day Parade, without using a trailer.\nPlan: \"Research different types of floats that are seen in the Macy's Day Parade to see if any are used without a trailer.\"\nGoal: Order food in a restaurant, where you don't speak the native language, without using a translator app.\nPlan: \"The plan would involve using gesture and pointing at items on the menu to describe what you would like the order. You could also find someone who speaks your language and persuade them to order/translate for you.\" [...] \nGoal: Jump over a six foot tall man, without using a trampoline.\nPlan: \"Fashion a catapult using a resistance band and a tree trunk, and launch yourself over the man.\"\nGoal: Build a bookshelf, without using wood.\nPlan: \"You need to go and purchase cinder blocks and sheets of plastic. Stack the cinder blocks as a base. You should do 3 rows for stability. You can then use the plastic sheets for shelves and the backing.\"\nGoal: Stop your canoe from falling down the waterfall, without using a paddle.\nPlan: \"1. Survey area for other things to grab onto . 2. Use hands in the water as makeshift paddle to maneuver your way to the possible objects (rocks, tree branches, etc). 3. Reach out to and grab onto said objects.\"[...]\nGoal: Fix a flat tire, without using a spare tire.\nPlan: \"Call roadside assistance or someone you know and have them bring a tire that fits you car. Second option is to call a towing company and have your car towed to a tire shop.\"\nGoal: Escape quicksand, without using a branch.\nPlan:\"I would",
\end{spverbatim}

Note, human-generated plans are couched in quotations and always end with a punctuation mark; periods were added to any human-generated plan that ended with a letter instead. New lines demarcated these \textit{goal:plan} pairs. We therefore queried GPT-3 by starting on an open parenthesis and used ``\texttt{." \textbackslash n}'' as our stop token. 

A comparable process was employed for the explanation domain, with the modification that the number of seed prompts was set to $N=12$ due to context window constraints (explanation stimuli tended to be longer than the goals used in the planning domain). Moreover, GPT-3 was instead seeded with pairs of the form: ``Scenario:'' and ``Explanation: This could have happened because''.

GPT-3 was prompted using a temperature of $0.5$ and queried for a maximum of $300$ tokens per \textit{rollout}. We define ``rollout'' as a single forward generation from the LLM. We set the maximum number of tokens to be greater than the most tokens used by any human participant to ensure that GPT-3 was given a fair chance to generate a sufficiently long plan or explanation. 

We collected $30$ plans and explanations per linguistic prompt. We manually removed rollouts which included derogatory language and regenerated as needed to ensure we had $30$ generations per prompt. Generations which did not end on a period, but reached the maximum number of tokens were spliced to end on the last period. If no period was generated in the rollout, the generation was discarded and re-generated.

As there is the possibility that such parsing may have impacted the semantics -- in addition to broadly degenerate, un-humanlike behavior potentially impacting ratings -- we reasoned that our \textit{prescreen} stage would naturally ensure that only the most human-like language was maintained in order to give GPT-3 the fairest chance possible in our later evaluations.

\subsection{Statistical analyses}

We include the full syntax for all linear regression models employed to run the statistical tests conducted in Part I: 

\begin{itemize}
    \item Within-group (within humans, and within LLM) sensitivity to constraints: \texttt{score $\sim$ condition + (1 | rater\_id) + (1 | prompt)}
    \item Between-group (humans vs. LLM) sensitivity to constraints: \texttt{score $\sim$ (source * condition) + source + condition + (1 | rater\_id) + (1 | prompt)}
    \item Robustness to prompt typicality: \texttt{score $\sim$ (source * typicality\_score) + source   +  typicality\_score + (1 | rater\_id) + (1 | prompt)}
\end{itemize}

\section{S2. Part II Supplemental Details} \label{sec:s_part_ii}

We next clarify the stimuli used, models compared, and evaluation set-up employed in \textit{Part II: Integrating language with structured reasoning models.} Associated code can be found under the \href{https://github.com/collinskatie/structured_flexible_and_robust/tree/main/Part_II}{``Part\_II'' directory} of our repository. 

\subsection{Set-up and stimuli creation} We design a problem setting to mimic the planning domain of Part I; however, here, we need to design tasks such that we can exactly verify whether a plan successfully solved a goal. The open-ended nature of the planning problems of Part I prohibits this degree of control, which is a necessary step to design a solid testbed to compare models. To that end, we construct a synthetic grammar -- over both \textit{goals and actions}  -- that can be directly mapped into formal predicates for goals, and PDDL actions. This enables us to \textit{execute} generated plans and evaluate success, as discussed in the section entitled ``Plan simulation enviornment''. 

Possible actions in our grammar include: \textit{stack, unstack}, and \textit{stackfromtable}. Example initial configurations and goals took the form of those shown in Fig. 4 of the main text. Configurations entail stacking problems, where each stacking problem includes a random set of $N=4$ items selected from a pre-defined vocabulary of everyday household items (e.g., ``plate'', ``keyboard''). We generate $100$ test configurations using this grammar, and for each configuration, build three increasingly constrained settings -- as discussed in Part II. Constraints are formed by sampling one of the constraints that fully specifies the goal condition and adding this to the goal. For every object mentioned in additional constraints (we disclude the first constraint), we swap the object with an out-of-distribution object (where out-of-distribution is defined with respect to what may not usually be found on a household table; e.g., ``meteorite'', ``corduroy pants'') to inject a dimension of atypicality to mirror stimuli used in Part I.

A listing of all stimuli used can be found in \href{https://github.com/collinskatie/structured_flexible_and_robust/tree/main/Part_II/dataset/simple_stack/problems}{our repository}. 

\subsection{Prompting}

We next discuss how we prompt the LLMs used in this section. We employ an LLM-as-Planner (GPT-Neo \cite{gpt-neo} which generates plans in natural language (NL) to directly mirror the LLM set-up used in Part I. Additionally, P+S relies on an LLM (Codex \cite{codex}) to parse the initial configurations and goal specification into formal predicates. 

\paragraph{LLM-as-Planner} For the \textit{LLM-as-Planner model} (a vanilla LLM prompted to produce an entire plan, consisting of a sequence of actions, given a linguistic goal), we construct a few-shot prompt analogous to those in \textit{Part I} consisting of a header with a set of ``training" example (\textit{goal}, \textit{plan}) pairs separated by the same delimiter, and then ending in the desired \textit{goal} for which the LLM should produce a plan. For all goals, we use a header consisting of the same sequence of n=3 (\textit{goal}, \textit{plan}) examples (which are disjoint from the goals we evaluate.
 A sample prompt containing these training example used can be found in the  \href{https://github.com/collinskatie/structured_flexible_and_robust/tree/main/Part_II#readme}{README} within the \href{https://github.com/collinskatie/structured_flexible_and_robust/tree/main/Part_II}{``Part\_II'' directory}. Solutions are structured to follow ``Actions:'', and ``Initially:'' is used as the stop token, as it would indicate the start of a new planning problem. Generations were run with a temperature of $0.05$. 

\paragraph{P+S prompting} The LLM-model used in the P+S model is only used as a "parser": it transduces linguistic goals into a symbolic program containing the formal environment predicates for this goal.

Therefore, the LLM in this example is prompted with a header with a set of ``training" example (\textit{goal}, \textit{parsed\_goal\_program\_predicate}) pairs separated by the same delimiter, and then ending in the desired \textit{goal} for which the LLM should produce a parse. For all goals, we use a header consisting of the same sequence of n=3 (\textit{goal}, \textit{parsed\_goal\_program\_predicate}) examples, which are drawn from the same set of training examples as in the LLM-as-planner baseline prompt (but these are shown only with goal parses, not complete plans).

An open parentheses cues Codex to start generating PDDL, and ";" is used as a stop token. A sample prompt can be found in the same \href{https://github.com/collinskatie/structured_flexible_and_robust/tree/main/Part_II#readme}{README}. 

\subsection{Plan simulation environment}\label{sec:plan_sim}

Our set-up consists of an LLM-as-Planner generating plans in natural language, and our P+S model generating plans as programs (e.g., PDDL). We therefore needed to design a scheme to compare such natural language against plans -- namely, our ``Plan simulation enviornment''. Our set-up automatically parses LLM-generated language into a program using our synthetic grammar. Both programs are then evaluated as to whether they achieved the goal. Planning success is coded in a binary (1 = success, 0 = fail) fashion. If the goal state is not reached, then the plan is deemed to have failed.

\subsection{Statistical analyses}

The statistical analyses run in Part II used the following syntax:  

\begin{itemize}
    \item Between-group (LLM-as-Planner vs. P+S) performance: \texttt{succeed $\sim$ method + (1 | id)}
    \item Between-group (LLM-as-Planner vs. P+S) sensitivity to constraints: \texttt{succeed $\sim$ (method * constraints) + method + constraints + (1 | id)}
\end{itemize}

We conducted an initial analysis (\texttt{succeed $\sim$ constraints + (1 | id)}) to see if both models were impacted by constraints, akin to Part I; we find that the LLM-as-Planner and P+S are indeed both impacted by constraints ($p < 0.001$). 

\subsection{Qualitative analysis of failure modes}

We found that the LLM-as-Planner was able to generate valid actions, but not consistently valid plans. For instance, when prompted with the following initial configuration and goal:

\begin{spverbatim}
Initially:
The writing pad rests on the table.
The notebook is on the writing pad.
The tissue box is on the notebook.
There is nothing on the tissue box.
The tablet rests on the table.
There is nothing on the tablet.
Goal:
There is nothing on the notebook.
\end{spverbatim}

The LLM generated the plan: \textit{Move the tablet onto the notebook.}, a direct violation of the goal constraint; therefore, an invalid plan. 

\bibliographystyle{apacite}

\setlength{\bibleftmargin}{.125in}
\setlength{\bibindent}{-\bibleftmargin}

\bibliography{supplement}